\documentclass[letterpaper, 10pt, conference]{ieeeconf}

\usepackage{graphicx}
\usepackage{cite}
\usepackage{microtype}
\usepackage{amsmath, amsfonts}
\usepackage{booktabs}

\graphicspath{{figs/}}

\newcommand{\ignore}[1]{%
}

\IEEEoverridecommandlockouts 

\title{\LARGE \bf
Method for Robotic Motion Compensation during PET Imaging of Mobile Subjects
}

\author{Junxiang Wang$^{1}$, Iulian Iordachita$^{1}$, Peter Kazanzides$^{2}$%
\thanks{$^{1}$Dept. of Mechanical Engineering, Johns Hopkins University, Baltimore, MD 21218, USA {\tt\small (email: [jwang334, iordachita]@jhu.edu)}}%
\thanks{$^{2}$Dept. of Computer Science, Johns Hopkins University, Baltimore, MD 21218, USA {\tt\small (email: pkaz@jhu.edu)}}%
}

\begin{document}
    
\maketitle

\begin{abstract}
  Studies of the human brain during natural activities, such as locomotion, would benefit from
  the ability to image deep brain structures during these activities.
  While Positron Emission Tomography (PET) can image these structures,
  the bulk and weight of current scanners
  are not compatible with the desire for a wearable device. This has motivated the design
  of a robotic system to support a PET imaging system around the subject's head and to move the
  system to accommodate natural motion.
  We report here the design and experimental evaluation of a prototype robotic system that senses
  motion of a subject's head, using parallel string encoders connected between the robot-supported imaging
  ring and a helmet worn by the subject.
  This measurement is used to robotically move the imaging ring (coarse motion correction) and to
  compensate for residual motion during image reconstruction (fine motion correction).
  Minimization of latency and measurement error are the key design goals, respectively, for coarse and
  fine motion correction.
  The system is evaluated
  using recorded human head motions during locomotion, with a mock imaging system consisting of
  lasers and cameras, and is shown to provide an overall system latency
  of about 80\,ms, which is sufficient for coarse motion correction and collision avoidance,
  as well as a measurement accuracy of about 0.5\,mm for fine motion correction.
\end{abstract}

\section{Introduction}

The physical distance between humans and robots has decreased significantly since robots were
introduced in factories in the 1960s, isolated from humans by gates and other safety barriers.
In the 1990s, robotic surgery brought robots into contact with humans, and more recently, \emph{collaborative}
robots have been introduced to enable humans and robots to safely work together.
The distance between humans and robots decreases even further with prosthetics and exoskeletons,
where the robot is physically attached to the human to replace a missing limb or to provide additional support.

In this work, we explore a different paradigm for integration of humans and robots, which is to
introduce a robot that maintains close proximity without physically contacting the human.
Our specific goal is to position a PET imaging ring around the head of a subject and accurately measure the
position of the head with respect to the imaging ring. The measurement is required to enable the robot to move
the imaging ring to compensate for limited motion of the subject (coarse motion correction) and for motion
correction in the image reconstruction algorithm (fine motion correction). The concept for robotic PET imaging
was introduced by Majewski et al. \cite{Majewski2020} based on prior experience with
a wearable imager, called Helmet PET \cite{Majewski2011}, that was limited by the unavoidable trade-off between
weight and sensitivity.

We previously analyzed motion capture and accelerometer data from subjects during overground and treadmill walking,
respectively, to determine the typical range of head motion, as well as head velocity and acceleration \cite{Liu2021}.
We assumed a helmet to be worn by the subject with dimension 263\,mm x 215\,mm, which corresponds to a minimal
(radial) clearance of about 18\,mm between the helmet and an imaging ring of 300\,mm diameter.
Our simulation results indicated that it is feasible for a robot to compensate for the head motion,
without colliding with the imaging ring, given a system latency of less than 100\,ms \cite{Liu2021}.

We subsequently developed a measurement system that consists of six string encoders connected between the
PET imaging ring and a helmet worn by the subject \cite{Wang2022}. The encoders were arranged in a parallel
configuration (Stewart Platform) and experimentally shown to provide a static measurement accuracy of about 0.5\,mm,
which matches the accuracy desired by the PET imaging scientists to correct for motion artifact during image reconstruction.
In \cite{Wang2022}, we also surveyed existing proximity sensing techniques used in various fields and showed that our system
provides higher accuracy, higher robustness, and lower latency.

\begin{figure}[!b]
  \centering
  \includegraphics[width=\linewidth]{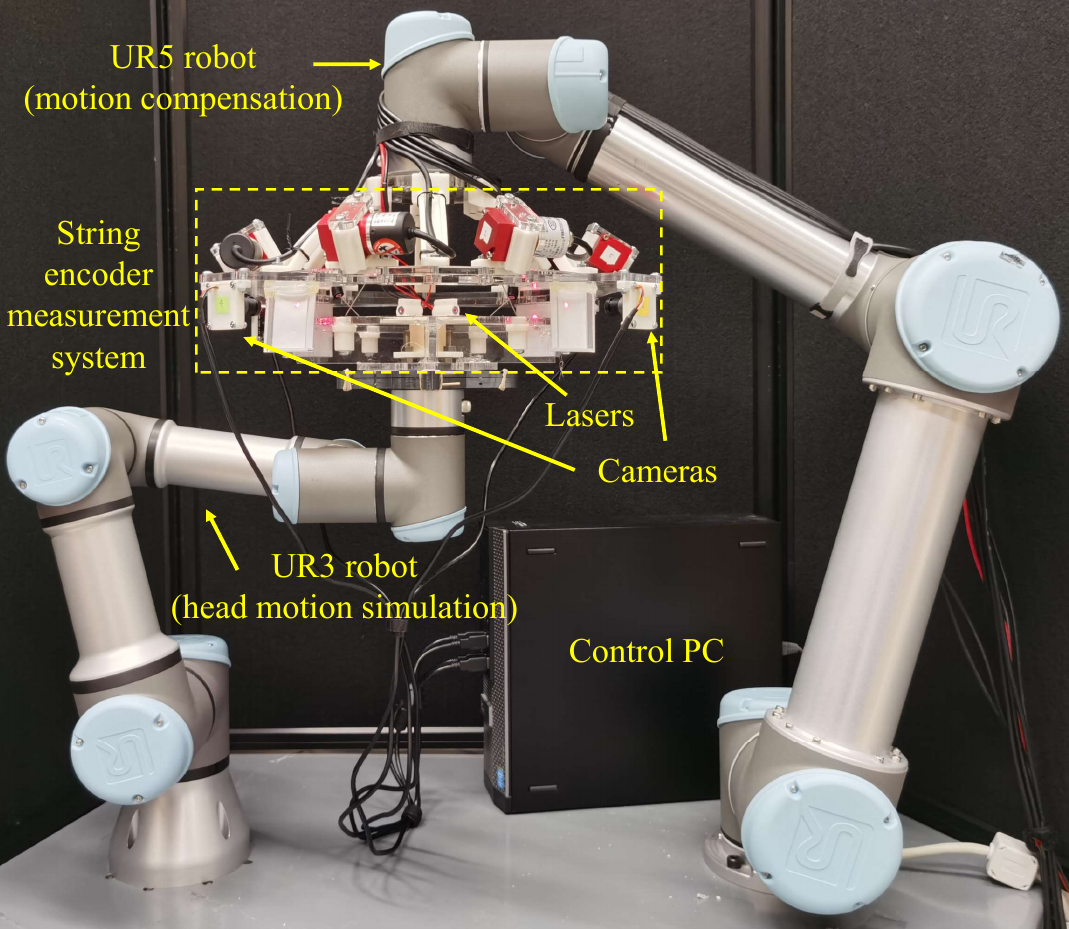}
  \caption{Overall setup with UR3 and UR5 robots, the string encoder system, as well as the mock PET imaging system.}
  \label{fig:system-pic}
\end{figure}

Our goal in this work is to extend this static measurement into dynamic conditions, where a low-latency
robotic system is designed around the position and velocity measurements of the string encoder system.
The test input is provided by a second robotic system that reproduces the recorded human head motion,
as shown in Figure \ref{fig:system-pic}.
This proof-of-concept robotic system will enable us to experimentally verify the feasibility of robotic
compensation for human head motion.

Our proposed system involves human-robot interaction, where the robot must safely move in proximity of the human,
and thus prior work includes studies of robot safety and especially methods for
enhancing safety through predictive control. Mainprice \cite{Mainprice2013} developed a prediction framework in
which a library of human motion data is captured in a model with regression performed offline, and subsequently
real-time prediction can be made given some actual perceived human motion, accompanied by appropriate robot
motion planning and execution. Koppula suggested a similar approach for anticipation \cite{Koppula2016}.

Haddadin's works on collision detection, evaluation, and reaction \cite{Haddadin2017} \cite{Haddadin2009},
as well as Dogramadzi's methods of safety assessment, especially for robot hazard and error \cite{Dogramadzi2021},
are relevant to the risk assessment and safety design of this type of system.
However, this paper focuses on the situation where the
robotic system functions normally, and collision is avoided through measurement
and compensation of motion, whereas future safety design will focus on ensuring correct
system operation, including redundant motion sensing, and evaluating the risk due to failure.

The following sections present the design of the proposed system for supporting
the PET imaging ring around the subject's head (Section~\ref{sec:system-design}) and the
associated control algorithm (Section~\ref{sec:ur5-control}). We then explain our experimental setup with
a robotic system for emulating human head motion (Section~\ref{sec:ur3-control}) and an optical system for
emulating PET imaging (Section~\ref{sec:laser-camera}). The results include the quality of the encoder-based
motion measurement (Section~\ref{sec:string-results}), the fidelity with which human head motion is reproduced
(Section~\ref{sec:ur3-result}), and the performance of both coarse and fine motion corrections (Sections~\ref{sec:motion-comp} and~\ref{sec:img-recon}).

\section{System}
\label{sec:system}

This section describes the components of our prototype robotic motion measurement and compensation
system with string encoders and a UR5 robot, shown with other experiment components in Fig.~\ref{fig:system-pic}.
We first discuss the overall system structure that focuses on latency minimization, followed by details
of the UR5 robot controller and its utilization of the string encoder measurements.

\subsection{Prototype Robotic PET System Design}
\label{sec:system-design}

\begin{figure}[tbh]
  \centering
  \includegraphics[width=\linewidth]{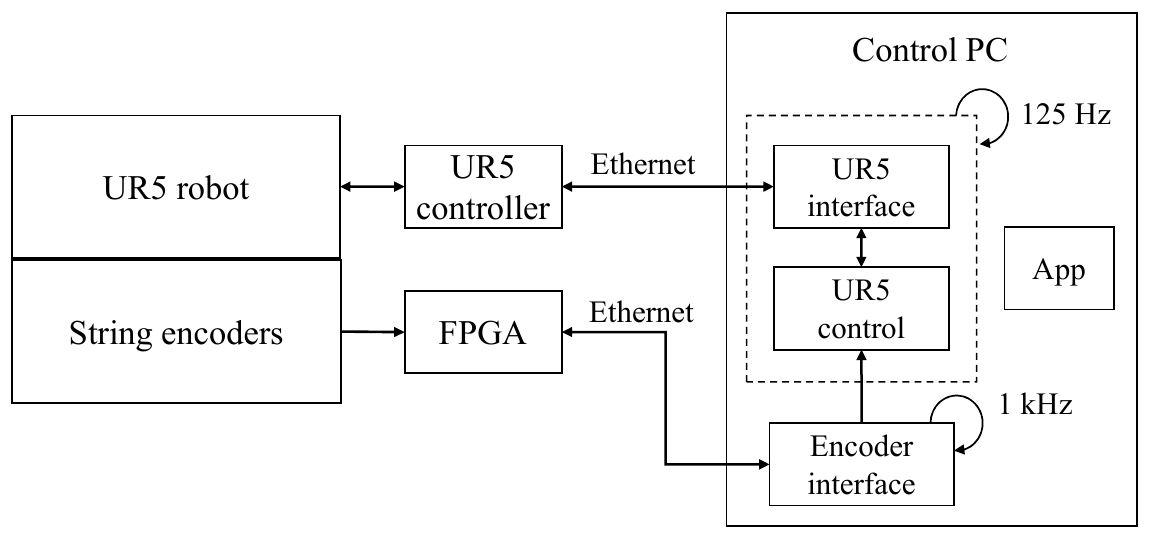}
  \caption{System block diagram, showing string encoders to measure position and velocity of helmet
  (worn by subject, not shown) with respect to imaging ring supported by UR5 robot.}
  \label{fig:system-block}
\end{figure}

A block diagram of the prototype system for tracking and compensating for human head motion during PET imaging
is shown in Fig.~\ref{fig:system-block}. A key design goal was to minimize latency, which led to the use
of a Field Programmable Gate Array (FPGA) for processing the encoder signals and a
component-based software architecture where all components are in a single process on the Control PC.

Head motion tracking is performed by six string encoders\footnote{MPS-XXXS-200MM-P, Miran Industries, China.
Three units had their rotary encoder replaced with E30S4-1000-3-T-5, Autonics, Korea due to hardware defects.},
with resolution of 60 quadrature counts per millimeter,
connected in parallel between a head-mounted helmet and a (mock) PET imaging ring and able to measure a
6 degree-of-freedom (DOF) Cartesian pose, as studied in \cite{Wang2022}.
The encoder signals are processed by an FPGA with a custom interface board. The FPGA board is an open
source design \cite{FPGA} (Rev 2.1) that is primarily used for the da Vinci Research Kit (dVRK) \cite{Kazanzides2014}.
The custom interface board consists of connectors and receivers that convert the 5V encoder signals to
the 3.3V level expected by the FPGA. The FPGA firmware is a customized version of the standard
dVRK firmware \cite{firmware} (Rev 7), with the following modifications:

\begin{itemize}
\item Provides up to 8 channels of encoder feedback, rather than just 4 channels.
\item Saves the encoder counter value when an index pulse occurs, and provides this value to the host PC.
\end{itemize}

The encoder feedback consists of position, obtained by quadrature counting of encoder pulses, and
velocity, which is based on the measured time between encoder edges. In particular, the velocity
measurement has been optimized to provide accurate estimates with low latency, even for low-resolution
encoders \cite{Wu2018}.
The FPGA board is connected to the host PC via Ethernet and uses the UDP protocol for
low-latency data exchange.

The software is based on the open-source \emph{cisst} libraries \cite{cisst} that
support components with separate threads in a single process, with efficient (lock-free)
data exchange mechanisms. In addition to an application component (App) that provides a
user interface for exercising the system and collecting data, there are three primary components:

\textbf{Encoder Interface:} This is a periodic component that reads the string encoder positions and velocities
from the FPGA via the UDP interface and then performs the forward kinematic computations to convert
them to Cartesian positions and velocities. This component is executed at 1\,kHz.

\textbf{UR5 Interface:} This is an instance of the open-source sawUniversalRobot component \cite{sawUR},
which interfaces to the UR5 controller via its real-time script interface. The component thread
waits for a feedback packet from the UR5 controller, which is provided at 125\,Hz, and thus the
component executes at 125\,Hz.

\textbf{UR5 Control:} This component implements the robot motion controller described
in Section~\ref{sec:ur5-control}. It shares the same thread as the UR5 Interface component
and thus is executed synchronously with that component at 125\,Hz. In particular, the UR5 Interface
component transfers control to the UR5 Control component after receiving the UR5 feedback packet.
UR5 Control computes the commanded velocity, then returns control to UR5 Interface,
which sends the commanded velocity to the UR5 robot.
The UR5 robot
cannot handle the expected weight
of an actual PET imaging ring, estimated at 15-20\,kg, but the final system could scale up by using
a larger robot, such as the UR20 or a custom design.

\subsection{Robotic Motion Control System}
\label{sec:ur5-control}

\begin{figure}[tbh]
  \centering
  \includegraphics[width=\linewidth]{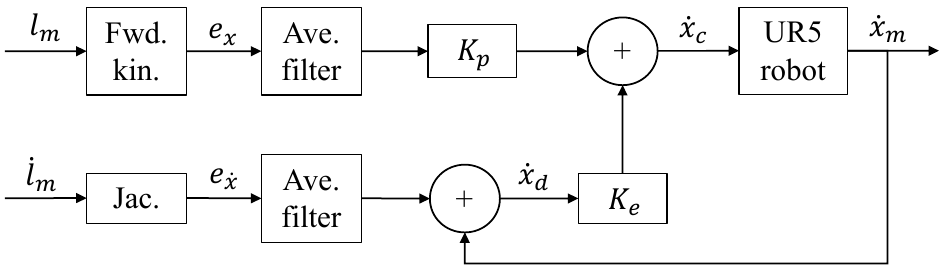}
  \caption{Block diagram of UR5 robot velocity controller, used to move imaging ring with respect to helmet.
    $\dot{x}_m$ is the measured UR5 robot Cartesian velocity.
    $l_m$ and $\dot{l}_m$ are the measured string encoder lengths (positions) and velocities, respectively,
    between imaging ring and helmet.}
  \label{fig:ur5-block}
\end{figure}

Using the 6 DOF position and velocity reported by the encoder interface, the UR5 robot moves itself in the
direction to restore the helmet ring and the imaging ring of the string encoder system back to its nominal
configuration. The robot is controlled in Cartesian velocity mode, via the real-time script interface, because this
has been shown to provide lower latency than position control \cite{Timm2015}. The general control logic for this
robotic motion compensation is velocity feedforward control with position correction (see Fig.~\ref{fig:ur5-block}).

The Encoder Interface directly outputs the string length $l_m$ and velocity $\dot{l}_m$
for each string encoder. As previously described \cite{Wang2022},
sending the set of $l_m$ through the parallel robot forward kinematics
gives the 6 DOF Cartesian pose measured by the system, and applying the Jacobian to $\dot{l}_m$
yields the Cartesian velocity. In our setup, these quantities respectively correspond to the position difference
$e_x$ and the velocity difference $e_{\dot{x}}$ between the head and the PET device. These are averaged
over 30 consecutive samples to reduce the noise inherent
in the velocity measurement. This filter window size was empirically determined to achieve the noise reduction needed for smooth robot movement,
while not inducing too much latency. All aforementioned operations are performed in the encoder interface at 1\,kHz,
which increases the measurement latency by about 15\,ms.

The position correction component of the motion controller is obtained by multiplying $e_x$ by a constant proportional gain $K_p$.
The desired feedforward velocity $\dot{x}_d$ is the sum of the measured UR5 robot Cartesian velocity $\dot{x}_m$
and the velocity difference $e_{\dot{x}}$. The translational (first three) and rotational (latter three) components of
$\dot{x}_d$ are each given a separate nonlinear, exponential gain, $K_e^T$ or $K_e^R$. The velocity
commanded to the UR5 robot $\dot{x}_c$ is the sum of the position difference and velocity terms, each
multiplied by their respective gains, as shown in Fig.~\ref{fig:ur5-block} and the equation below:
\begin{equation}
  \dot{x}_c = K_p e_x + \begin{bmatrix}
    K_e^T I_3 & 0_{3,3} \\ 0_{3,3} & K_e^R I_3
  \end{bmatrix} \dot{x}_d
\label{eq:ur5-vc}
\end{equation}
where $I_n$ denotes the $n$-dimensional identity matrix, and $0_{m,n}$ the $m$ by $n$ zero matrix.

The gain factor $K_e^T$ is a function of $\delta^T$, the magnitude of the translational
components of $e_x$, such that the gain approaches 0 when $\delta^T$ is small and approaches 1 when it is large. In
particular, its value is determined by the following piecewise continuous function:
\begin{equation}
  K_e^T(\delta^T) = \begin{cases}
    \frac{\delta^T}{\delta^T_{\text{max}}}e^{\frac{\delta^T}{\delta^T_{\text{max}}}-1} & \text{for } 0 \leq \delta^T < \delta^T_{\text{max}} \\
    1                                                                          & \text{for } \delta^T \geq \delta^T_{\text{max}}
  \end{cases}
\end{equation}
where $\delta^T_{\text{max}}$ is a threshold beyond which $K_e^T$ should always evaluate to 1, and is set as 20\,mm in our
implementation. $K_e^R$ is defined analogously, with $\delta^R$ being the ``angle'' value in the axis-angle representation
of the rotational component of $e_x$, and with $\delta^R_{\text{max}}$ as 14 deg.
This nonlinear gain improves the performance of the system by avoiding oscillations around the target pose due
to noisy velocity measurements.
Moreover, $\delta_{\text{max}}$'s were empirically determined such
that the actual $\delta$'s would rarely reach the maximum levels under typical motions, so only partial velocity is fed
forward, which reduces overshoot due to the averaging filter phase lag.

\section{Experiments}

This section first discusses our experimental setup for simulation of human head motion and PET imaging, followed by
descriptions of experiments for evaluating the system's performance for both coarse and fine motion correction.

\subsection{Robotic System for Simulating Human Head Motion}
\label{sec:ur3-control}

\begin{figure}[!t]
  \centering
  \includegraphics[width=\linewidth]{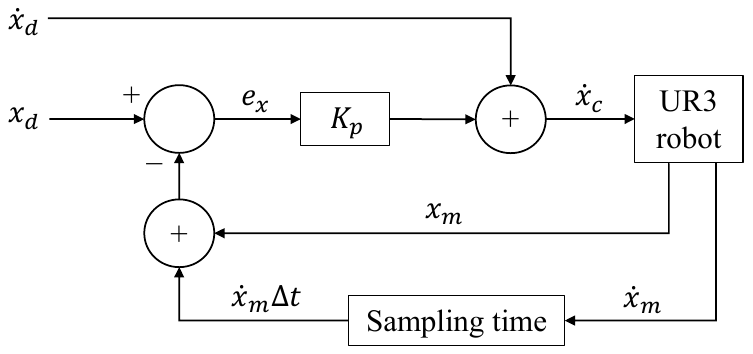}
  \caption{Block diagram of UR3 robot velocity controller, used to emulate human head (helmet) motion.
    $x_d$ and $\dot{x}_d$ are the desired helmet Cartesian position and velocity, respectively,
    obtained from recorded human head motion data. $x_m$ and $\dot{x}_m$ are the measured
    Cartesian position and velocity, respectively, of the UR3 robot.}
  \label{fig:ur3-block}
\end{figure}

To reproduce the collected human head motion from our previous experiments \cite{Liu2021}, we use a UR3
robot. We re-computed 6 DOF head motion trajectories from the original optical marker data collected at 60\,Hz, using
a total of five marker positions: $P_{LEYE}$, left side of the front of the head; $P_{REYE}$,
right side of the front of the head; $P_{LEAR}$, left side of the back of the head; $P_{REAR}$, right side of the back
of the head; and $P_{C7}$, cervical vertebra. The coordinate system for these positions is defined such that $+x$ is in
the direction of locomotion of the subjects, and $+z$ is vertically upwards. A linear regression is performed on the
$P_{C7}$ time series to estimate the position of the body center as the subjects walked overground in a straight line,
denoted as $P_{C7, fit}$. We compute the translation of the head center with: \begin{equation}
  t_c = \frac{1}{4}(P_{LEYE} + P_{REYE} + P_{LEAR} + P_{REAR}) - P_{C7, fit}
\label{eq:tc}
\end{equation}
The first term in equation (\ref{eq:tc}) finds the mean of the four markers on the head to approximate the position
of the head center, and the subtraction of $P_{C7, fit}$ is to remove the effect of overground walking to
simulate treadmill walking. We denote the orientation of the head by $R_c = [\textbf{r}_x,\textbf{r}_y,\textbf{r}_z]$.
Since $P_{LEYE}$ and $P_{REYE}$ lie on approximately the same horizontal level, with $P_{LEAR}$ and $P_{REAR}$ on a
different level, these four points roughly form a common plane.
$\textbf{r}_z$ is then determined as the norm vector of the best-fit plane
with a positive z-component. $\textbf{r}_x$ is given by the unit vector in the direction:
\begin{equation}
  (P_{LEYE} + P_{REYE}) - (P_{LEAR} + P_{REAR}).
\end{equation}
Lastly, $\textbf{r}_y$ is simply calculated with $\textbf{r}_z \times \textbf{r}_x$. Since these vectors likely do not
align with the principal axes of the head geometry, the inverse of the starting rotation is applied to each rotation
in the trajectory as an approach at normalization. The rotations are
parametrized with Euler ZYX convention and, combined with translations, encode each trajectory as a sequence of
6-vectors $x_d$. A Savitzky-Golay filter with fourth degree polynomial fitting over a frame length of 17 is applied to
each sequence for smoothening and calculation of the first and second derivatives---Cartesian velocities $\dot{x}_d$ and
accelerations $\ddot{x}_d$.
$x_d$ and $\dot{x}_d$ are then provided as input to the UR3 motion controller shown in Fig.~\ref{fig:ur3-block}.

At each time step (every 16.7\,ms), the velocity commanded to the UR3 is the sum of $\dot{x}_d$
and a correction term based on the pose error
between $x_d$ and the UR3's predicted pose at the next time step, calculated from
its measured 6 DOF pose $x_m$ and velocity $\dot{x}_m$. This pose difference $e_x$ is multiplied by a
constant gain $K_p$ and added to $\dot{x}_d$ to form the total commanded velocity to the UR3, $\dot{x}_c$.

To statistically confirm the fidelity of reproducing human head motion,
we recorded $x_m$ and $\dot{x}_m$ from the robot and performed a first-order Savitzky-Golay
differentiation on $\dot{x}_m$ to obtain the actual acceleration $\ddot{x}_m$.
Section~\ref{sec:ur3-result} compares the means and standard deviations of these three actual quantities
to their corresponding desired values.

The UR3 robot integrates into the system described in Section~\ref{sec:system} by attaching the helmet
ring of the string encoder system onto the UR3's end effector, as shown in Fig.~\ref{fig:system-pic}.
This motion reproduction system can be reused even when the UR5 robot
is replaced by a robot capable of delivering the larger payload (15-20\,kg) that
will be required to support an actual PET imaging system. 

\subsection{Optical System for Simulating PET Imaging}
\label{sec:laser-camera}

\begin{figure}[tbh]
  \centering
  \includegraphics[width=\linewidth]{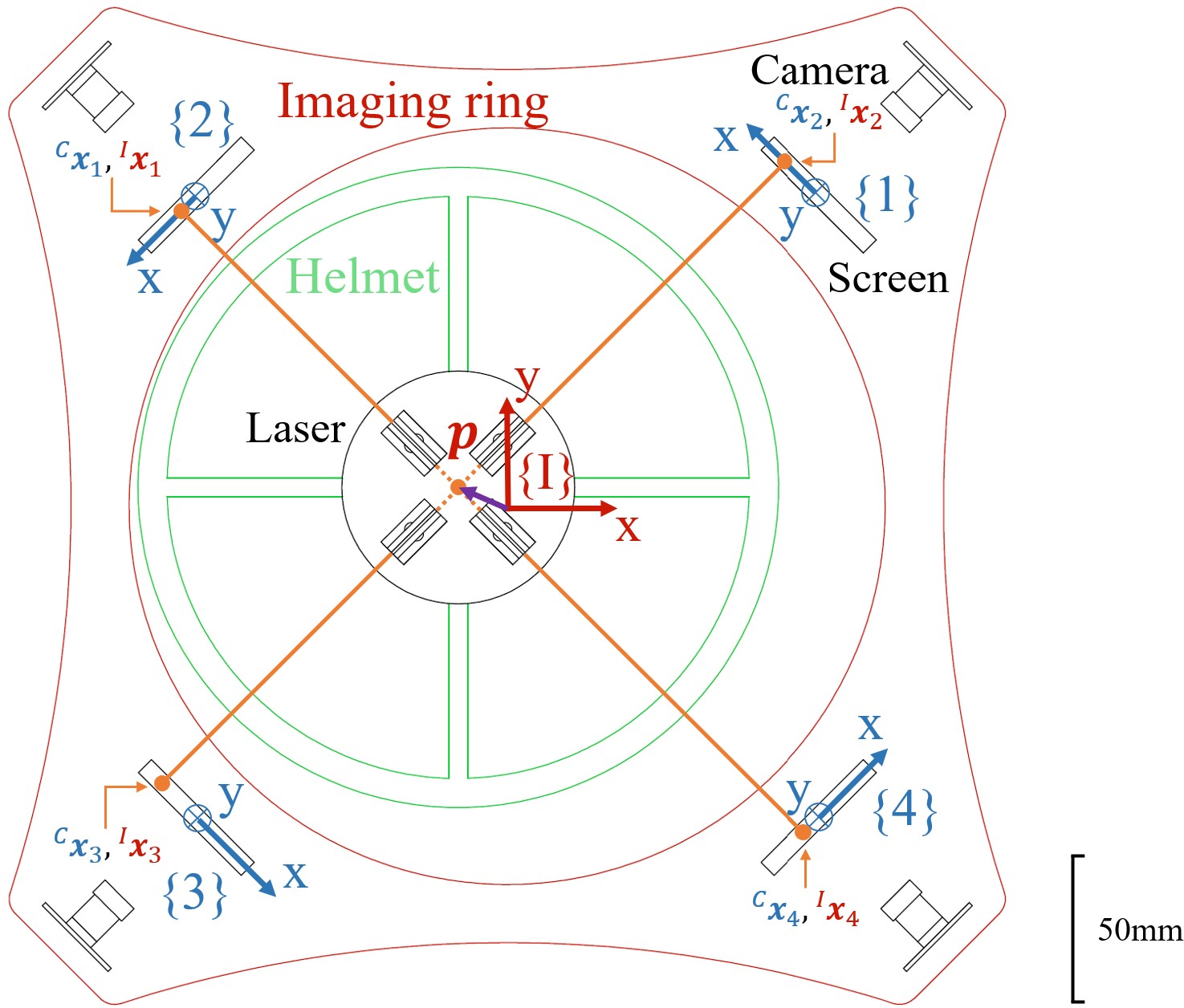}
  \caption{Top view diagram showing outlines of prototype helmet (green) and imaging ring (red) with mock
           PET imaging system (lasers and cameras).}
  \label{fig:top-diagram}
\end{figure}

\begin{figure}[tbh]
  \centering
  \includegraphics[width=\linewidth]{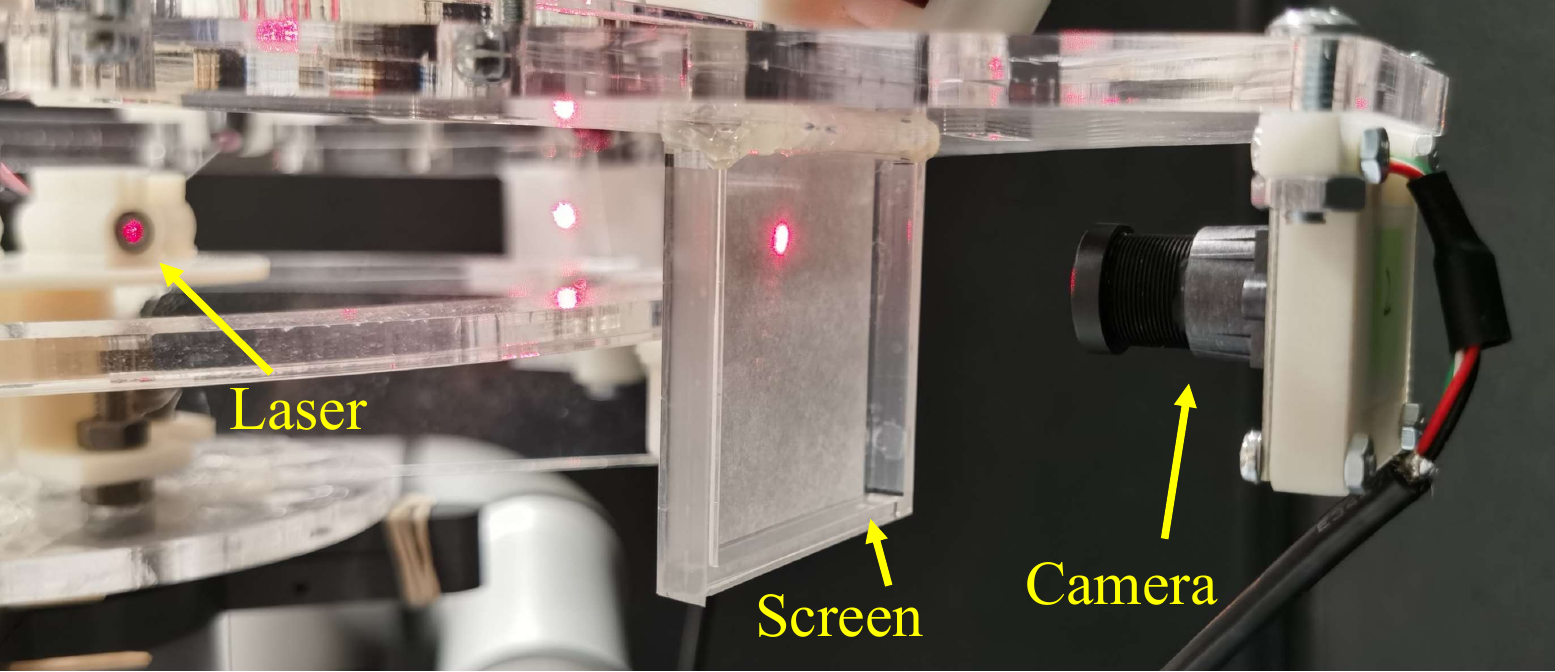}
  \caption{Close-up view of laser, screen, and camera.}
  \label{fig:laser-camera}
\end{figure}

Since the ultimate goal of our robotic system is to provide not only coarse motion correction with robot movement but also
fine motion correction with image reconstruction, we simulated the radioactive PET imaging process with a simplified optical setup. 
Specifically,
since PET relies on the emission and detection of positrons shooting in opposite directions from tissues, we used backward-facing laser diodes to simulate pairs of positrons, and cameras with screens in front of them to simulate PET detectors.
We employed four laser diodes (VLM-650-03 LPA, Quarton Inc., Korea) and four cameras (UB0234, Arducam, China).
The lasers were
mounted to the center of the helmet ring, facing outward horizontally, spaced apart evenly by 90 deg, and the cameras were mounted to the outer
edge of the imaging ring, facing inward to align with the laser directions in the nominal configuration of the string encoder system.
An acrylic frame supporting a screen of white paper is placed between each camera and laser to block the laser beam, with
the paper thin enough for the camera on the other side to clearly capture the laser dot.
Fig.~\ref{fig:top-diagram} shows a top-view diagram of the lasers and cameras,
Fig.~\ref{fig:system-pic} shows their location within the overall system, and Fig.~\ref{fig:laser-camera} shows one active laser diode, its corresponding dot on the paper screen, and a camera.

The two pairs of laser beams emulate two orthogonal lines emanating from a single radioactive point source in the brain.
Our reconstruction algorithm determines the position of the point source $\textbf{p}$ in the imaging ring frame $I$, analogous
to actual PET image reconstruction. We first locate the position of the laser dot on each
camera screen, yielding four 2D coordinates $\{^C\textbf{x}_i = (x_i,y_i)\}_{i=1}^4$ with $(0,0)$ being the position in the nominal
configuration of the string encoder system.
The 3D representations of $^C\textbf{x}_i$ in frame $I$, $^I\textbf{x}_i$, are then computed. $\textbf{p}$ is determined as the intersection
of two lines: one connecting $^I\textbf{x}_1$ and $^I\textbf{x}_3$, the other one connecting $^I\textbf{x}_2$ and $^I\textbf{x}_4$.
Fig.~\ref{fig:top-diagram} shows the numbering and coordinate system orientations, as well as a hypothetical non-nominal configuration
where there is a non-zero translation in the x-y plane between the imaging ring and the helmet. Labeled in the figure are the points
where the lasers would hit the screens, as well as the computed position of the point source $\textbf{p}$, shown as a purple vector near the center.
The outer diameter of the helmet ring is drawn scaled-down in order to exaggerate the allowable translation for
the benefit of clearer illustration.
The figure is a simplified scenario, and in general this method for determining $\textbf{p}$ can be used for arbitrary 6 DOF pose of
the helmet.

\subsection{Experimental Procedures}

We first evaluated the string encoder system's measurement accuracy and latency during dynamic motions by commanding the UR3 with a 2\,Hz sinusoidal wave, which is a frequency typical to natural human head
motion based on our previous data\cite{Liu2021}. We evaluated the measurements against the motion provided by the UR3; the other components
of our system were inactive for this experiment.

We then evaluated the performance of the complete system using a real human head motion trajectory. With all previously described
components and controllers in effect, we commanded the UR3 with one of the processed head motion trajectories. Starting with comparison
between the recorded motion and the UR3 reproduced motion as described in Section~\ref{sec:ur3-control}, we then evaluated the
performance of both the coarse motion correction and the fine motion correction. In the former, the UR5 motion was compared against
the UR3 motion with error and latency assessments. In the latter, the string encoder measurements were used to adjust the images
captured by the cameras to account for residual motion (not compensated by the robot), and in the ideal case, the laser intersection would be seen as stationary
after this reconstruction process.

\section{Results}
\label{sec:results}

\subsection{Head Measurement Accuracy and Latency}
\label{sec:string-results}

\begin{figure}
  \centering
  \includegraphics[width=\linewidth]{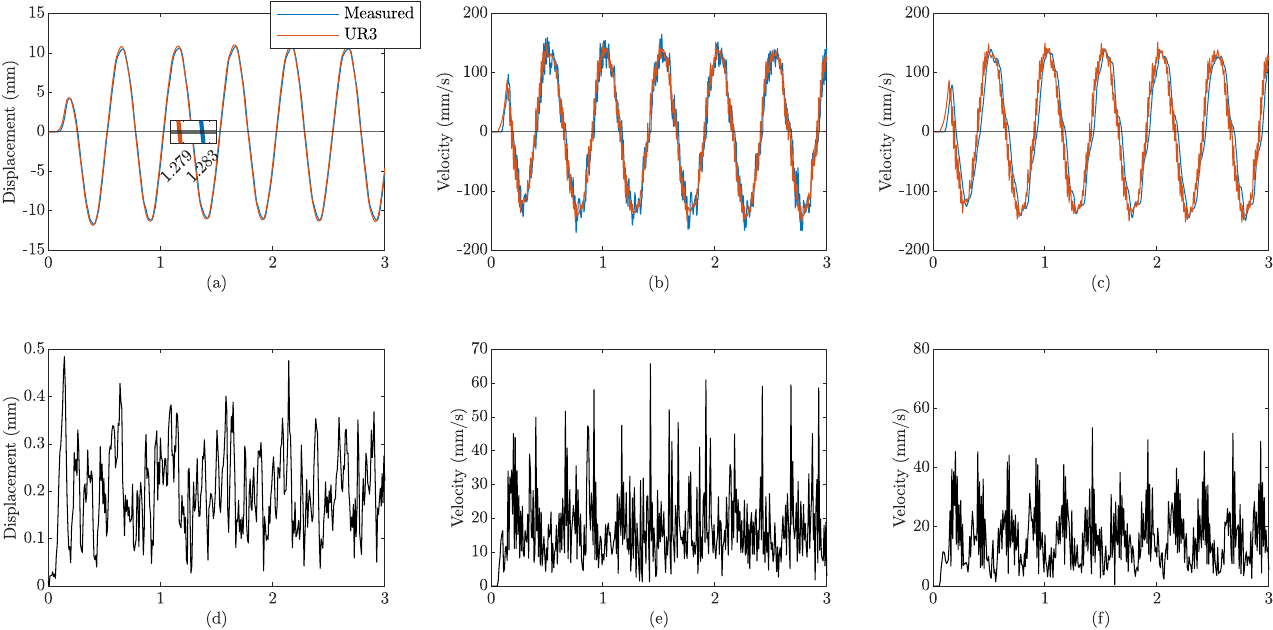}
  \caption{Dynamic position and velocity measurements from string encoder system against actual UR3 motion.}
  \label{fig:encoder-ur3-pos}
\end{figure}

Fig.~\ref{fig:encoder-ur3-pos}(a) shows the displacement of the UR3 during the 2\,Hz sine wave, compared against the string encoder system's measured displacement.
A zoomed-in view is provided in the center to illustrate the very low latency of around 4\,ms between the two. Fig.~\ref{fig:encoder-ur3-pos}(d) shows their
difference after correction for this latency, which signify the measurement accuracy of the string encoder system; this has a mean of 0.19\,mm with a standard deviation of 0.09\,mm.
Fig.~\ref{fig:encoder-ur3-pos}(b) and Fig.~\ref{fig:encoder-ur3-pos}(e) are the corresponding plots for velocity, and the measurement error has a mean of 12\,mm/s with 
a standard deviation of 13\,mm/s. The larger amount of noise in velocity measurements also confirms the decision to apply filtering, the result of which is shown in
Fig.~\ref{fig:encoder-ur3-pos}(c). Note that the difference in Fig.~\ref{fig:encoder-ur3-pos}(f) is computed after correction for both measurement latency and filter latency.
The filtered velocity measurement accuracy has a mean of 11\,mm/s with a standard deviation of 12\,mm/s. These results demonstrate that the high static position measurement
accuracy shown previously \cite{Wang2022} extends to dynamic measurements.

\subsection{Accuracy of Reproducing Human Head Motion}
\label{sec:ur3-result}

\begin{figure}[!b]
  \centering
  \includegraphics[width=\linewidth]{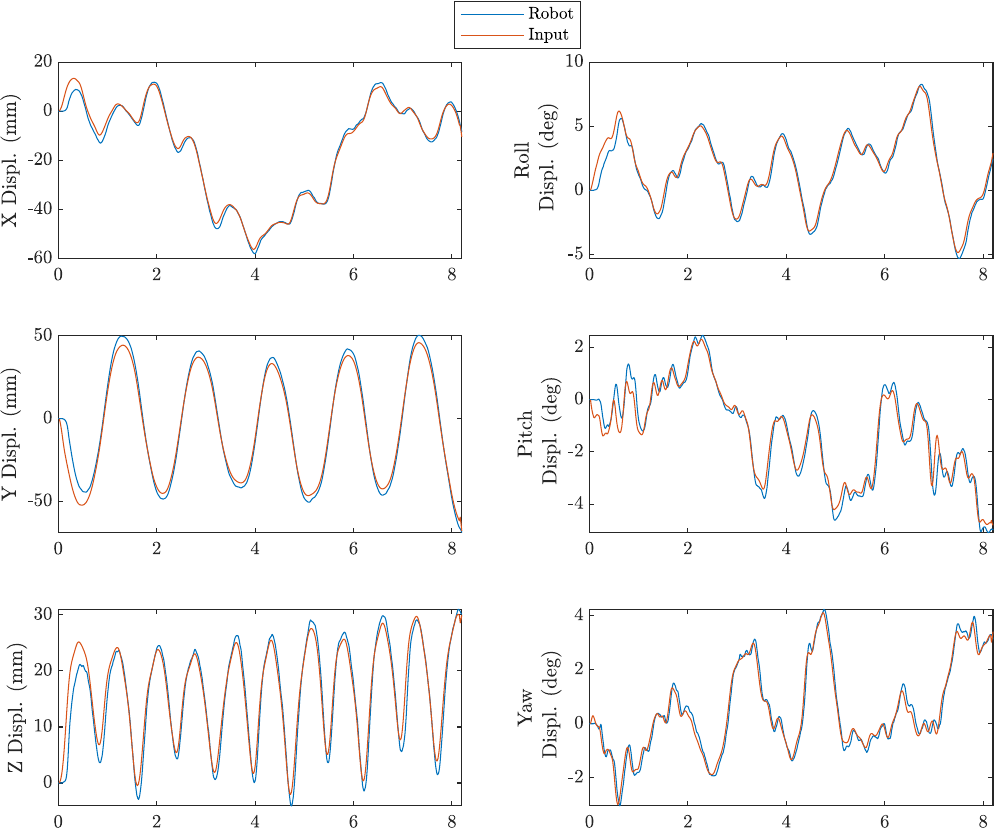}
  \caption{Response of UR3 in reproducing human head motion in all three axes. Horizontal axes represent time in seconds.}
  \label{fig:ur3-head}
\end{figure}

The UR3 reproduction of human head motion is evaluated against the input head trajectory visually in Fig.~\ref{fig:ur3-head}
and statistically in Table~\ref{tab:stats}. As shown in the figure, after a slight lag
in the beginning, the UR3 is able to execute the commanded motion with high accuracy. Since our goal
with the UR3 is only for it to accurately emulate human head motion, any lag does not affect the
overall motion compensation system, as long as its motion is consistent with the input.
The statistics in the table, which do not include the first 1\,s of motion,
confirm that the UR3 output motion follows the characteristics of the desired input motion.

\begin{table}
  \centering
  \caption{Mean and standard deviation of position, velocity, and acceleration data for both desired (D)
    and measured (M) motion. Units are in mm, mm/s, and mm/s\textsuperscript{2}}
  \setlength{\aboverulesep}{.2ex}
  \setlength{\belowrulesep}{.35ex}
  \begin{tabular}{crrrrrr}
    \toprule
                          & \multicolumn{2}{c}{\textbf{Displ.}}  & \multicolumn{2}{c}{\textbf{Vel.}} & \multicolumn{2}{c}{\textbf{Acc.}} \\
    \cmidrule(lr){2-3} \cmidrule(lr){4-5} \cmidrule(lr){6-7}
                          & \multicolumn{1}{c}{\textbf{Mean}} & \multicolumn{1}{c}{\textbf{S.D.}} & \multicolumn{1}{c}{\textbf{Mean}} & \multicolumn{1}{c}{\textbf{S.D.}} & \multicolumn{1}{c}{\textbf{Mean}} & \multicolumn{1}{c}{\textbf{S.D.}}          \\
    \midrule
    \textbf{X, D}        & -15.0  & 20.2    & -6.72     & 141      & 45.5      & 815                    \\
    \textbf{X, M}        & -15.6  & 20.2    & -8.90      & 130      & -0.99      & 705                    \\
    \cmidrule(lr){1-7}
    \textbf{Y, D}        & -6.14  & 32.4    & -2.39      & 69.5      & 12.7      & 439                    \\
    \textbf{Y, M}        & -4.36  & 43.1    & -1.41      & 67.5      & 10.9      & 493                    \\
    \cmidrule(lr){1-7}
    \textbf{Z, D}        & 16.6  & 8.33   & 3.08      & 78.3      & -18.3      & 767                  \\
    \textbf{Z, M}        & 15.8  & 9.27    & 3.58      & 76.8      & -7.73      & 844                   \\
    \cmidrule(lr){1-7}
    \textbf{Roll, D}        & 1.78  & 2.76    & 0.51      & 13.1      & -2.59      & 116                    \\
    \textbf{Roll, M}        & 1.63  & 2.84    & 0.39      & 12.8      & 1.94      & 150                    \\
    \cmidrule(lr){1-7}
    \textbf{Pitch, D}        & -1.31  & 1.71    & -0.27     & 10.57     & 3.70      & 138                   \\
    \textbf{Pitch, M}        & -1.28  & 1.85    & -0.72      & 9.72      & 0.40      & 202                    \\
    \cmidrule(lr){1-7}
    \textbf{Yaw, D}        & 0.53  & 1.62    & 0.25      & 8.52      & -1.82      & 103                    \\
    \textbf{Yaw, M}        & 0.54  & 1.67    & 0.44      & 8.97      & 1.30     & 171                    \\
    \bottomrule
  \end{tabular}
  \label{tab:stats}
\end{table}

In addition, the UR3 head motion controller serves as a representation of the open-loop response of the UR5 motion
compensation controller---in the ideal situation without measurement noise, the UR5 controller would function identically
to the UR3 controller. Hence, although latency is not a concern for the UR3 controller, we
measured its lag with a ramp test signal of 80\,mm/s velocity up to a distance of 200\,mm to serve as a reference for the UR5 behavior.
A numerical analysis of the collected data indicates that 
the robot is able to achieve a mean latency of about 16\,ms over the test trajectory.

\subsection{Coarse Motion Correction}
\label{sec:motion-comp}

\begin{figure}[tbh]
  \centering
  \includegraphics[width=\linewidth]{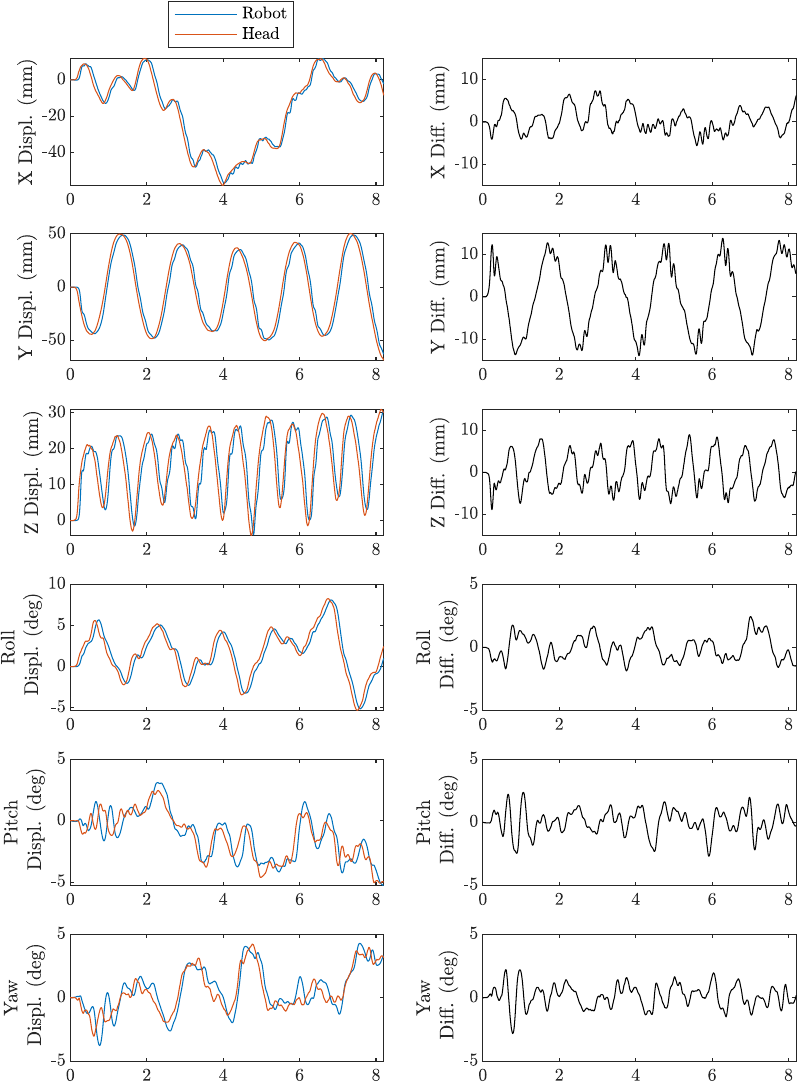}
  \caption{Comparison between UR5 (robot) position and UR3 (head) position for recorded head motion, with actual motion
           trajectories (left) and differences between the two (right). Horizontal axes represent time in seconds.}
  \label{fig:ur5-ur3}
\end{figure}

As the UR3 executed the head trajectory discussed in the previous section, the UR5 performed motion compensation.
The trajectories of both robots are depicted in Fig.~\ref{fig:ur5-ur3}. Shown on the left are the measured displacements for all six axes collected from
the two robots, and on the right are the differences between the trajectories. As shown in the first column, the tracking
performance for translations and rotation about the z-axis (roll) is better than for rotations about the x- and y-axes
(yaw and pitch). One explanation is that the range of motion in pitch and yaw is the smallest, and due to the averaging
filter, the robot would not compensate for such small motions. In addition, our previous evaluation of the
string encoder system \cite{Wang2022} showed that there are slightly higher measurement inaccuracies for pitch and yaw.
Nonetheless, the second column of Fig.~\ref{fig:ur5-ur3} does not show significant discrepancy across the trajectory differences in all 
rotation axes.

For the entire travel, the norm of translational tracking error in the x-y plane has a maximum of 14.5\,mm and a mean of 5.8\,mm.
The error in the z-direction is less relevant as our proposed 18\,mm clearance between the helmet and the imaging ring is within
the x-y plane. Nonetheless, even considering the z-error, the maximum is 15.7\,mm with a mean of 6.6\,mm, which is still within the stated
clearance. The rotational tracking error has a maximum of 3.8\,deg and a mean of 1.1\,deg. Since this rotation is measured with
respect to the head center, it would not contribute to the available clearance if we assume the head being spherical.

Furthermore, the UR5 is able to maintain a typical lag of about 55\,ms (and up to 80\,ms in some cases) while compensating
for the head motion. The typical lag is approximately the sum
of the string encoder measurement latency discussed in Section~\ref{sec:string-results} (4\,ms), the filtering latency described
in Section~\ref{sec:ur5-control} (15\,ms), and the robot motion latency measured in Section~\ref{sec:ur3-result} (16\,ms).

\subsection{Fine Motion Correction}
\label{sec:img-recon}

\begin{figure}
  \centering
  \includegraphics[width=\linewidth]{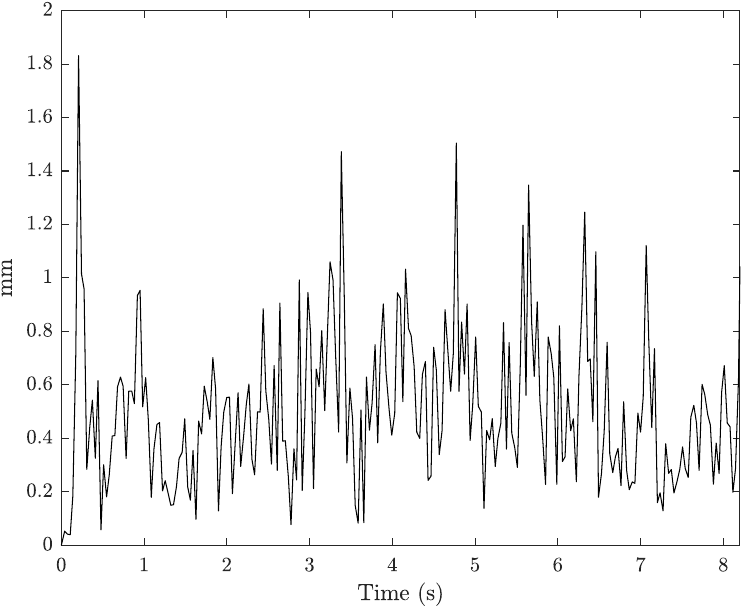}
  \caption{Norm of uncorrected distance after fine motion correction.}
  \label{fig:cam-err}
\end{figure}

After applying fine motion correction with the string encoder measurements, any nonzero position $\textbf{p}$ of the laser intersection
relative to its starting position as seen from the cameras should ideally be corrected to zero. The norms of these
position vectors after correction are plotted in Fig.~\ref{fig:cam-err}.

These uncorrected distances have a mean of 0.51\,mm and a standard deviation of 0.28\,mm,
which is in line with the desired 0.5\,mm accuracy.
Part of the inaccuracy originates
from the string encoder system, which we previously evaluated to have an accuracy of less than 0.3\,mm for a $\pm10$\,mm
range of translational motion \cite{Wang2022}.
The cameras also possess inaccuracy in localizing the red dot, and in our setup, each pixel
approximately corresponds to 0.1\,mm of distance. In addition, the camera exposure time for a single frame is around
15\,ms, hence images captured during motion show a trail of red light instead of a single dot. We tried to identify
the end of the trail and relate that position to the time the image was taken for synchronization purposes. However, this
approach becomes more complicated in regions where there is a direction change in the relative motion between the
helmet and imaging ring. Therefore, some of the inaccuracy is due to our camera setup, rather than the proposed
robotic system.

\section{Conclusions}
\label{sec:conclusions}

We presented the design of a prototype robotic system that maintains a specified position offset with respect to
a human head, based on measurements from string encoders connected in parallel between the robot and a helmet attached to the head.
The intended use is for PET imaging during natural activities, such as walking, to support neuroscience research,
but the concept could be applied to other scenarios where it is desirable to have a robot system follow the motion
of a human without significant physical contact. Technically, the string encoders create a small amount of physical
contact between the robot and human, and the spring-loaded tensioning of the strings causes a small force
to be exerted on the head. This disadvantage, however, is offset by the high reliability of the mechanical connection and the
lower latency and higher accuracy that can be achieved compared to non-contact alternatives such as proximity
sensing and optical tracking. In this application, accurate measurement of the residual motion 
is important because it enables motion correction during PET image reconstruction.

Our results demonstrate that a UR5 robot, with hardware and software designed to minimize latency
to less than 100\,ms, can provide adequate motion correction. Specifically, the system can keep residual 
motion within the designed 18\,mm clearance between the PET imaging ring and helmet \cite{Liu2021} and the
string encoder system's dynamic accuracy enables sub-millimeter motion correction during PET image reconstruction.
The final design, however, will require a robot that can support the higher weight of a PET imaging ring (15-20\,kg) and it is desirable
for the string encoder measurement system accuracy to be better than 0.5\,mm.

Future work includes improvements to the measurement accuracy, as discussed above.
In addition, the presented robot controller is purely reactive, based on measured
head position and velocity, and performance could be improved by adopting predictive control \cite{Koppula2016}.
Our recorded data indicates that human head motions are highly periodic during locomotion,
which suggests the feasibility of this approach.
Finally, development of a system for neuroscience research with human subjects will require a significant
effort in risk analysis and safety design, as well as evaluation of the human trust of, and comfort with, a robot
system in close physical proximity.

\section*{Acknowledgements}
Stan Majewski initiated the investigation of robotic compensation for head motion during PET imaging.
The recorded head motion data was provided by Caroline Paquette.
Ti Wu provided the kinematic design of the string encoder system.

\bibliographystyle{IEEEtran}
\bibliography{IEEEabrv, references}

\end{document}